# Detecting PTSD in Clinical Interviews: A Comparative Analysis of NLP Methods and Large Language Models

Feng Chen[1,2], Dror Ben-Zeev[2], Gillian Sparks[2], Arya Kadakia[2], Trevor Cohen[1,2]

[1]*Department of Biomedical Informatics and Health Education, University of Washington, Box 358047 Seattle, WA 98195, USA*

[2]*Behavioral Research in Technology and Engineering (BRiTE) Center, Department of Psychiatry and Behavioral Sciences, University of Washington, 3751 W Stevens Wy NE Seattle, WA 98195, USA*

*Email: fengc9@uw.edu, dbenzeev@uw.edu, gsparks@uw.edu, arya1kadakia@gmail.com, cohenta@uw.edu*

Post-Traumatic Stress Disorder (PTSD) remains under-detected in clinical settings, presenting opportunities for automated detection to identify at-risk patients. This study evaluates natural language processing approaches for binary PTSD classification from clinical interview transcripts using the DAIC-WOZ dataset, which contains semi-structured interviews with standardized psychological assessments. We compared embedding-based methods (SentenceBERT/LLaMA with logistic regression), general and mental health-specific transformer models (BERT/RoBERTa), and large language model prompting strategies (zero-shot/few-shot/chain-of-thought). SentenceBERT embeddings with logistic regression achieved the highest overall performance (AUPRC=0.758±0.128), outperforming domain-specific end-to-end fine-tuning models like Mental-RoBERTa (AUPRC=0.675±0.084 vs. RoBERTa-base 0.599±0.145). Few-shot prompting using DSM-5 criteria and two examples yielded competitive results (AUPRC=0.737). Performance varied significantly across symptom severity and comorbidity status with depression, with higher accuracy for severe PTSD cases and patients with comorbid depression. Our findings highlight the potential of embedding-based methods and LLMs for scalable screening while underscoring the need for improved detection of nuanced presentations.

## 1. Introduction

Post-Traumatic Stress Disorder (PTSD) affects approximately 6% of the U.S. population, with significantly higher rates among veterans and trauma survivors.[1] Despite its prevalence, PTSD remains under-detected in primary care settings, with studies suggesting that around 30% of cases go unrecognized.[2] Traditional screening approaches rely on structured clinical interviews and self-report measures, which require substantial clinical expertise and patient engagement. The development of automated screening tools could significantly improve detection rates, particularly in resource-constrained settings where mental health specialists are limited.

Recent advances in natural language processing (NLP) and large language models (LLMs) offer promising avenues for mental health assessment through analysis of patient language.[3, 4] While numerous studies have explored computational approaches to mental health detection, most have focused on depression and anxiety, with relatively less attention paid to PTSD.[4, 5] Furthermore, existing PTSD detection research has predominantly relied on social media data or surveys rather than clinical interview, limiting applicability at the point of care.[6] Prior work in clinical NLP has explored various approaches including transformer-based models like BERT and RoBERTa, embedding-based methods using SentenceBERT, and prompting strategies with large language models. Traditional transformer models trained on general text corpora which may have limited capability to capture the nuanced linguistic markers of PTSD, such as trauma-specific disfluencies, avoidance semantics, or fragmented narrative coherence. Though domain-adapted variants like Mental-BERT show promise for depression classification, their efficacy for PTSD detection remains unproven.[7] Embedding-based approaches using SentenceBERT have demonstrated effectiveness for depression detection tasks.[8] while instruction-tuned large language models offer potential for clinical assessment through prompting strategies that leverage clinical knowledge without requiring extensive labeled training data.[9, 10] However, these approaches have rarely been applied to PTSD detection in clinical interview settings, representing a significant gap in the literature.

The Distress Analysis Interview Corpus - Wizard of Oz (DAIC-WOZ) dataset provides a unique opportunity to develop and evaluate computational methods for PTSD detection in a simulated clinical context.[11-13] This dataset contains semi-structured clinical interviews conducted by a virtual interviewer with standardized psychological assessments as ground truth. While previous studies using this dataset have primarily focused on depression detection or multimodal approaches combining audio, visual, and linguistic features, the potential of advanced NLP techniques specifically for PTSD detection remains underexplored.[14]

To address this gap, this study bridges critical gaps in computational PTSD detection through a multifaceted NLP framework that advances both methodological innovation and clinical relevance. Our research investigates three key aspects: First, we examine the efficacy of various language representation approaches for PTSD detection with models fine-tuned on labeled data, comparing general versus domain-specific pre-trained transformer models (BERT and RoBERTa compared with Mental-BERT and Menta-RoBERTa) with embedding-based methods using more recent architectures (SentenceBERT and LLaMA). This comparison aims to identify whether domain adaptation or architectural differences have greater impact on PTSD classification performance. Second, we explore different LLM prompting strategies (zero-shot, few-shot, and chain-of-thought) for PTSD classification, which to our knowledge represents the first application of instruction-tuned LLMs to PTSD detection in clinical interviews. This approach investigates whether LLMs can leverage clinical knowledge encoded during pre-training without requiring extensive labeled data. By comprehensively comparing these approaches, we contribute benchmark results for future research on automated PTSD detection using transcripts from clinical interviews. Third, we analyze how model predictions vary across symptom severity levels and comorbidity with depression, seeking to understand if computational approaches face similar challenges to human clinicians in distinguishing PTSD from related conditions, and in detecting subclinical or borderline cases.

This work has implications for the development of automated PTSD screening tools in mental health settings. By leveraging advanced NLP techniques for PTSD detection, our research may help address the significant gap in PTSD screening and assessment, particularly in primary care and other settings where specialized mental health resources are limited.

## 2. Methods

### 2.1. *Dataset*

For this study, we utilized the DAIC-WOZ dataset, a specialized collection of clinical interviews designed for psychological assessment research. The corpus contains semi-structured clinical interviews conducted by a virtual interviewer named Ellie, whose responses were controlled by human operators following a consistent protocol. The DAIC-WOZ dataset comprises multimodal data from 189 participants, including audio, video, and transcribed textual data. Each interview followed a clinical protocol designed to elicit information relevant to psychological assessment, beginning with neutral rapport-building questions before progressing to more specific inquiries about symptoms related to psychological distress. Pre-interview, participants completed standardized psychological assessments, including the PTSD Checklist-Civilian Version (PCL-C), which was used to establish ground truth diagnostic labels.[12] Among 189 participants, 56 (29.6% of the sample) met criteria for PTSD, while the remaining 133 were non-PTSD participants. Additionally, participants completed the Patient Health Questionnaire-8 (PHQ8) for depression assessment, enabling analysis of comorbid conditions. Among them, 42 (22.2% of the sample) were diagnosed as having depression and 147 were categorized as non-depressed participants.

### 2.2. *Data Preprocessing and Model Architecture*

This study focused on exploring the predictive power of natural language from clinical interviews. For our analysis, we focused exclusively on the patient-side transcripts, extracting only the verbal content produced by participants while excluding interviewer questions and prompts. This approach allowed us to isolate linguistic features directly attributable to the participants, minimizing potential confounding from interviewer speech patterns or question framing, which have been shown to influence model predictions of depression.[14]

For our baseline approach, we employed BERT (Bidirectional Encoder Representations from Transformers),[15] a pre-trained language model that generates contextual word embeddings capturing semantic information. We experimented with both a general domain “bert-base-uncased” model and a domain-specific "mental-bert" variant, that has undergone additional pre-training using mental health-related posts collected from Reddit.[7] To address BERT's 512-token input limitation with lengthy clinical transcripts, we developed a chunking strategy that segmented each transcript into 512-token portions. From each chunk, we extracted the representation from the [CLS] token, which provides a 768-dimensional vector serving as an aggregate representation of that segment. To create a fixed-length representation for the variable-length transcripts, we computed the mean of all chunk embeddings, thereby preserving information from the entire transcript while maintaining consistent dimensions for the downstream classifier. We also implemented RoBERTa, which was trained with a larger corpus than BERT and with augmented training objectives,[16] utilizing both "roberta-base" and "mental-roberta" variants.[7] Models were trained for 10 epochs using AdamW optimizer with weight decay regularization,[17] [18] with a learning rate of 2e−5, batch size of 4, and binary cross-entropy loss. To address class imbalance, we implemented weighted random sampling during training, assigning weights inversely proportional to class frequencies to ensure balanced representation of PTSD and non-PTSD cases in each mini-batch.[19] A linear layer with sigmoid activation was used as a classification layer for binary classification.

In addition to end-to-end fine-tuning approaches, we implemented embedding-based PTSD classification using representations generated from SentenceBERT and LLaMA followed by logistic regression. Embedding-based approaches produces semantically meaningful sentence embeddings directly without task-specific fine-tuning like BERT, making it computationally efficient for classification tasks. For SentenceBERT, we utilized the "paraphrase-multilingual-mpnet-base-v2" model, which produces 768-dimensional vectors for each text segment under 128 words.[20] Similar to our BERT implementation, we chunked transcripts and applied mean pooling to generate transcript-level embeddings. For LLaMA embeddings, we used pre-computed 16,384-dimensional vectors that were generated through mean pooling across entire transcripts without chunking, leveraging the model's ability to handle longer sequences effectively. Both embedding approaches were followed by logistic regression classification using scikit-learn's implementation (C=1.0, liblinear solver, maximum 1000 iterations). For the logistic regression models, balanced class weights were automatically computed as inversely proportional to class frequencies in the training data, effectively upweighting the minority PTSD class during optimization.

All models were evaluated using the same 5-fold stratified cross-validation to ensure robust and unbiased performance estimation across different data splits and across all methodologies.

### 2.3. *Large language models approach*

We then explored the use of Large Language Models (LLMs) for PTSD prediction through various prompting strategies. We used LLaMA 3.1 with 405B parameters, a state-of-the-art open weight large language model, for this task in comparison with the fine-tuned models.[21, 22] This approach allowed us to assess whether LLMs could effectively identify linguistic markers of PTSD in patient speech without fine-tuning on labeled data. For our zero-shot learning approach, we developed a clinically informed prompt that framed the task as a diagnostic assessment performed by a psychiatrist specializing in trauma. The prompt incorporates the DSM-5 diagnostic criteria for PTSD,[23] which cover four key symptom clusters: intrusion symptoms, avoidance, negative alterations in cognitions and mood, and alterations in arousal and reactivity. We excluded the PTSD diagnostic Criterion A (exposure to traumatic events) as our analysis focused on linguistic markers of current PTSD symptomatology that would be observable in the interview transcripts. This approach augments the LLM's ability to perform psychiatric assessment by providing explicit diagnostic criteria, without requiring examples from our dataset.

For few-shot learning, we enhanced the prompt by including a small number of labeled examples from our dataset. We selected the positive case with the highest PCL-C (PTSD Checklist – Civilian Version)[24] score and the first negative case with the lowest score we encountered in the dataset to serve as anchors for the classification task. These examples provided the model with concrete instances of how PTSD manifests linguistically in our specific clinical interview context.

We further incorporated a chain-of-thought methodology where the model was instructed to produce a step-by-step reasoning process before making the final classification. The prompt directs the model to first analyze the transcript for specific PTSD indicators across all symptom clusters, then synthesize this information into a brief rationale (under 100 words) justifying the classification decision, and finally output a binary classification (0 for no PTSD, 1 for PTSD). Our implementation followed a two-stage process: first, we used the chain-of-thought prompt to generate the detailed reasoning, then extracted this reasoning and fed it into a separate zero-shot classification prompt to obtain the final binary classification result and probability scores. Through this approach we aimed to improve classification accuracy by encouraging more structured analytical reasoning that mimics clinical diagnostic processes. The detailed prompt can be found in Table 1.

Table 1. Prompt Structure for PTSD prediction

| Component | Prompt |
|---|---|
| **Role specification** | You are a highly experienced psychiatrist specializing in trauma and mental health disorders. |
| **Task description** | Your task is to analyze patient transcripts—containing only the patient's speech—and classify whether the patient has PTSD or not. |
| **DSM-5 Criteria** | According to the DSM-5 diagnostic criteria, PTSD is characterized by: a. Intrusion Symptoms: At least one symptom such as recurrent, involuntary, and intrusive distressing memories of the traumatic event(s); recurrent distressing dreams related to the event(s); dissociative reactions (e.g., flashbacks) in which the event seems to recur; intense or prolonged psychological distress at exposure to internal or external cues that symbolize or resemble the traumatic event(s); or marked physiological reactions to such cues. b. Avoidance: Persistent avoidance of stimuli associated with the traumatic event(s), evidenced by efforts to avoid distressing memories, thoughts, or feelings about or closely associated with the event(s) and/or avoidance of external reminders (people, places, conversations, activities, objects, or situations) that trigger these memories. c. Negative Alterations in Cognitions and Mood: Two or more symptoms such as inability to remember an important aspect of the traumatic event(s) (typically due to dissociative amnesia); persistent and exaggerated negative beliefs or expectations about oneself, others, or the world; persistent, distorted cognitions about the cause or consequences of the traumatic event(s) leading to self-blame or blaming others; persistent negative emotional state (e.g., fear, horror, anger, guilt, or shame); markedly diminished interest in significant activities; feelings of detachment or estrangement from others; or a persistent inability to experience positive emotions. d. Alterations in Arousal and Reactivity: Two or more symptoms such as irritable behavior and angry outbursts (with little or no provocation); reckless or self-destructive behavior; hypervigilance; exaggerated startle response; problems with concentration; or sleep disturbances. |
| **Output format (Answer)** | Based on these criteria and your analysis of linguistic patterns, coherence, sentiment, and emotional expressions in the transcript, output 0 if there is no indication of PTSD and 1 if PTSD is present. Provide only the classification result (0 or 1) without any additional explanation. |
| **Output format (Answer + reasoning)** | Based on these criteria and your analysis of linguistic patterns, coherence, sentiment, and emotional expressions in the transcript, output 0 if there is no indication of PTSD and 1 if PTSD is present. Format Your Output as Follows: - The classification at the start in the format: "Final Classification: 0" or "Final Classification: 1". - Step-by-step reasoning for the classification in less than 100 words. |
| **Few-shot examples** | Here are two examples, one positive and one negative: Transcript for a participant with PTSD: xxx. Classification: 1; Transcript for a participant without PTSD: xxx. Classification: 0 |

## 2.4. *Performance Evaluation*

For each model, we report multiple complementary metrics to provide a comprehensive view of predictive performance. Area Under the Precision-Recall Curve (AUPRC) was selected as our primary metric due to its robust performance in imbalanced datasets, providing better discriminant ability for rare-case scenarios compared to AUROC.[25] AUPRC is particularly valuable in clinical settings where identifying positive cases (PTSD patients) is the primary concern, as it focuses on precision and recall trade-offs across different classification thresholds.

We also report Area Under the Receiver Operating Characteristic curve (AUROC) to evaluate discrimination performance across all possible classification thresholds, and balanced accuracy to account for potential class imbalance in our dataset. Balanced accuracy estimates the arithmetic mean of sensitivity and specificity, giving equal weight to performance on positive and negative classes regardless of their proportional representation. For all models, we determined optimal classification thresholds using the Equal Error Rate (EER) method, where the false positive rate equals the false negative rate. This threshold was then applied to calculate balanced accuracy, ensuring fair performance comparison across different model architectures.

For LLM-based prompting approaches, we implemented specialized methods to derive probability scores for metric calculation. For the zero-shot and few-shot approaches, we extracted logits from the model for both class labels (0 and 1) and converted to probabilities using SoftMax normalization. This approach provides a continuous measure of the model's confidence in each classification. For the chain-of-thought approach we employed a two-stage process, first collecting the reasoning output without the final classification, then feeding this reasoning back to the model to obtain logits for possible classifications. These logits were then normalized to probabilities for AUROC and AUPRC calculation. This methodology ensured comparable measurements across all model types despite their fundamental differences in classification approaches.

## 2.5. *Comorbidity and Severity Analysis*

To investigate the relationship between PTSD and depression diagnoses, we conducted a comprehensive stratified analysis by comorbidity status and symptom severity. For comorbidity analysis, we examined prediction performance separately for patients with and without comorbid depression, allowing us to assess whether models performed differently across these clinically distinct subgroups. This analysis was conducted using AUPRC as the primary metric to maintain consistency with our overall evaluation framework.

For severity analysis, we stratified participants into discrete PCL-C severity bins designed to have approximately equal sample sizes within each group, ensuring robust statistical comparisons across severity levels. For PTSD-positive participants, we created four severity bins: 39-46 (lowest severity), 47-53, 54-62, and 63-85 (highest severity). For PTSD-negative participants, we established four corresponding bins: 17-19 (lowest scores), 20-25, 26-30, and 31-56 (highest scores approaching clinical threshold). This overlap in score ranges between diagnostic groups, present in the original data, underscores the complexity of PTSD diagnosis and explains some of the classification challenges observed in our models. We calculated classification accuracy for each model within these severity bins to determine how symptom intensity affected model performance across the spectrum of PTSD presentations. The analysis was performed across all three approaches: end-to-end, embedding-based and prompt-based models, to identify consistent patterns and understand the challenges models face across different clinical presentations of PTSD.

## 3. Results

### 3.1. *Supervised Learning Approaches*

#### *3.1.1. End-to-End Fine-tuning Models*

The performance of different end-to-end fine-tuning BERT-based models for PTSD classification is presented in Table 2. Our analysis revealed substantial performance differences between models pre-trained on general text corpora versus those specialized for mental health domains.

The domain-specific models consistently outperformed their general-domain counterparts across all evaluation metrics. Mental-BERT achieved an AUPRC of 0.647±0.101, representing a 24.0% improvement over BERT-base (0.522±0.092). Similarly, Mental-RoBERTa demonstrated superior performance with an AUPRC of 0.675±0.084, a 12.7% increase from RoBERTa-base (0.599±0.145).

Mental-RoBERTa achieved the highest AUPRC performance (0.675±0.084) among all end-to-end fine-tuning approaches, establishing it as the best-performing transformer-based model for PTSD detection. Notably, while RoBERTa variants typically outperform BERT models in many natural language processing tasks due to their more robust pre-training, in our experiments the advantage of Mental-RoBERTa over Mental-BERT was modest: only outperformed in AUPRC but underperformed in AUC and balanced accuracy. The domain of pre-training emerged as the dominant factor influencing performance, with both mental health-specific models substantially outperforming their general-domain counterparts.

Table 2. Performance comparison of general and mental health-specific BERT and RoBERTa for PTSD detection

| | BERT-base | Mental-BERT | RoBERTa-base | Mental-RoBERTa |
|---|---|---|---|---|
| **AUPRC** | 0.522±0.0 92 | 0.647±0.101 | 0.599±0.145 | **0.675±0.084** |
| **AUC** | 0.723±0.108 | **0.794±0.059** | 0.723±0.107 | 0.786±0.046 |
| **Balanced Acc** | 0.702±0.101 | **0.739±0.091** | 0.662±0.088 | 0.695±0.092 |

#### *3.1.2. Embedding-based Models*

We then evaluated embedding-based classification using frozen pre-trained representations followed by logistic regression to assess whether this two-stage approach could outperform end-to-end fine-tuning within the supervised learning category.

The SentenceBERT embedding + LR approach achieved an AUPRC of 0.758±0.128, representing a 12.3% improvement over the best end-to-end model (Mental-RoBERTa: 0.675±0.084). This approach also demonstrated strong discriminative ability with an AUROC of 0.856±0.069 and balanced accuracy of 0.801±0.097. The LLaMA embedding + LR method also showed competitive performance with an AUPRC of 0.693±0.094, AUROC of 0.835±0.046, and balanced accuracy of 0.747±0.029, similarly exceeding Mental-RoBERTa's performance across all metrics (as shown in Table 3).

Both embedding-based approaches outperformed end-to-end fine-tuning, with SentenceBERT + LR demonstrating the strongest performance among supervised learning methods. This finding suggests that pre-trained sentence-level embeddings, when combined with logistic regression on a small dataset, can capture linguistic patterns associated with PTSD more effectively than end-to-end fine-tuning. The strong performance of both embedding approaches indicates that frozen language model representations contain rich semantic information that can be effectively leveraged by simpler linear classifiers. Moreover, the computational efficiency of embedding-based approaches makes them particularly attractive for clinical applications, as they eliminate the need for resource-intensive fine-tuning while delivering superior performance within the supervised learning category.

### 3.2. *Instruction-Tuned Language models*

We evaluated various prompting strategies for PTSD classification using LLaMA 3.1 to assess whether large language models could achieve competitive performance without requiring labeled training data or domain-specific fine-tuning.

Zero-shot classification using LLaMA prompted with DSM-5 criteria achieved an AUPRC of 0.701, AUROC of 0.841, and balanced accuracy of 0.751 (Table 3). This approach leveraged the model's pre-trained knowledge by providing explicit psychiatric diagnostic criteria in the prompt, demonstrating that LLMs can effectively utilize clinical guidelines to identify linguistic markers of PTSD without any training examples from our dataset.

Adding one positive and one negative example case to the prompt improved performance across most metrics. The few-shot approach achieved an AUPRC of 0.737, balanced accuracy of 0.804, and the highest AUROC of 0.875 among all instruction-tuned methods (Table 3). The superior AUROC performance indicates exceptional discriminative ability, suggesting that strategically selected examples enhance the model's ability to distinguish between PTSD and non-PTSD cases.

Chain-of-thought prompting, which required the model to provide step-by-step diagnostic reasoning before classification, achieved an AUPRC of 0.579, AUROC of 0.705, and balanced accuracy of 0.681 (Table 3). This approach showed the lowest performance among instruction-tuned methods, indicating that explicit diagnostic reasoning steps do not improve PTSD classification accuracy in this context.

### 3.3. *Cross-Method Performance*

Table 3 presents a comprehensive comparison across all approaches. SentenceBERT + LR achieved the highest AUPRC (0.758±0.128) among all methods, representing the best overall performance for PTSD detection. However, few-shot prompting demonstrated the best discriminative ability (AUROC: 0.875) and competitive balanced accuracy (0.804).

Within supervised learning approaches, embedding-based methods consistently outperformed end-to-end fine-tuning, with both SentenceBERT + LR and LLaMA + LR exceeding Mental-RoBERTa across all metrics. Among instruction-tuned approaches, the strong performance of zero-shot and few-shot approaches demonstrates that these models have successfully encoded substantial domain knowledge relevant to mental health assessment during pre-training, enabling competitive performance without labeled training data.

The results reveal distinct trade-offs between approach categories: supervised learning methods achieved the highest performance but require labeled training data, while instruction-tuned approaches achieved competitive performance with few trained data but need substantial computational resources to host the large models, offering advantages in scenarios where clinical training data is scarce or unavailable.

Table 3. Performance comparison of fine-tuning, embedding-based, and prompting approaches for PTSD detection

| | **End-to-end** | **Embedding-based** | | **Prompt-based** | | |
|---|---|---|---|---|---|---|
| | Mental RoBERTa | SBERT + LR | LLaMA + LR | LLaMA ZS | LLaMA FS | LLaMA CoT |
| **AUPRC** | 0.675±0.084 | **0.758±0.128** | 0.693±0.094 | 0.701 | 0.737 | 0.579 |
| **AUC** | 0.786±0.046 | 0.856±0.069 | 0.835±0.046 | 0.841 | **0.875** | 0.705 |
| **Balanced Acc** | 0.695±0.092 | 0.801±0.097 | 0.747±0.029 | 0.751 | **0.804** | 0.681 |

## 3.4. *Model Predictions Patterns: Severity and Comorbidity Effects*

### *3.4.1. Severity Analysis*

Results of our evaluation of model performance across different PTSD severity levels, based on PCL-C self-report scores, are shown in Figure 1. All six approaches (Mental-RoBERTa, SentenceBERT + LR, LLaMA + LR, and the three LLaMA prompting methods) exhibit similar performance patterns across PTSD severity levels, suggesting consistent challenges in classification based on symptom intensity.

For PTSD-positive participants, most models demonstrated improved accuracy with increasing severity. Several approaches achieved perfect or near-perfect accuracy for the most severe cases (63-85 range): LLaMA + LR (100%), few-shot (100%), Mental-RoBERTa (92.9%), zero-shot (92.9%), and chain-of-thought (92.9%), while SentenceBERT + LR achieved 85.7%. These findings suggest that extreme cases at both ends of the severity spectrum (very low or very high PCL-C scores) are easier for models to classify correctly, while cases with moderate severity present the greatest challenge.

Conversely, for PTSD-negative participants, all models showed declining accuracy with increasing severity scores, indicating that PTSD-negative participants with more severe symptoms are more likely to be inaccurately classified as having PTSD. Accuracy dropped substantially from the lowest severity bin (17-19) to the highest bin (31-56) across all approaches: Mental-RoBERTa (80.6% to 46.9%), SentenceBERT + LR (83.3% to 59.4%), LLaMA + LR (86.1% to 56.2%), zero-shot (88.9% to 59.4%), few-shot (94.4% to 62.5%), and chain-of-thought (72.2% to 56.2%). This suggests that subclinical cases approaching the diagnostic threshold pose greater classification challenges, likely due to their linguistic similarities with mild PTSD presentations.

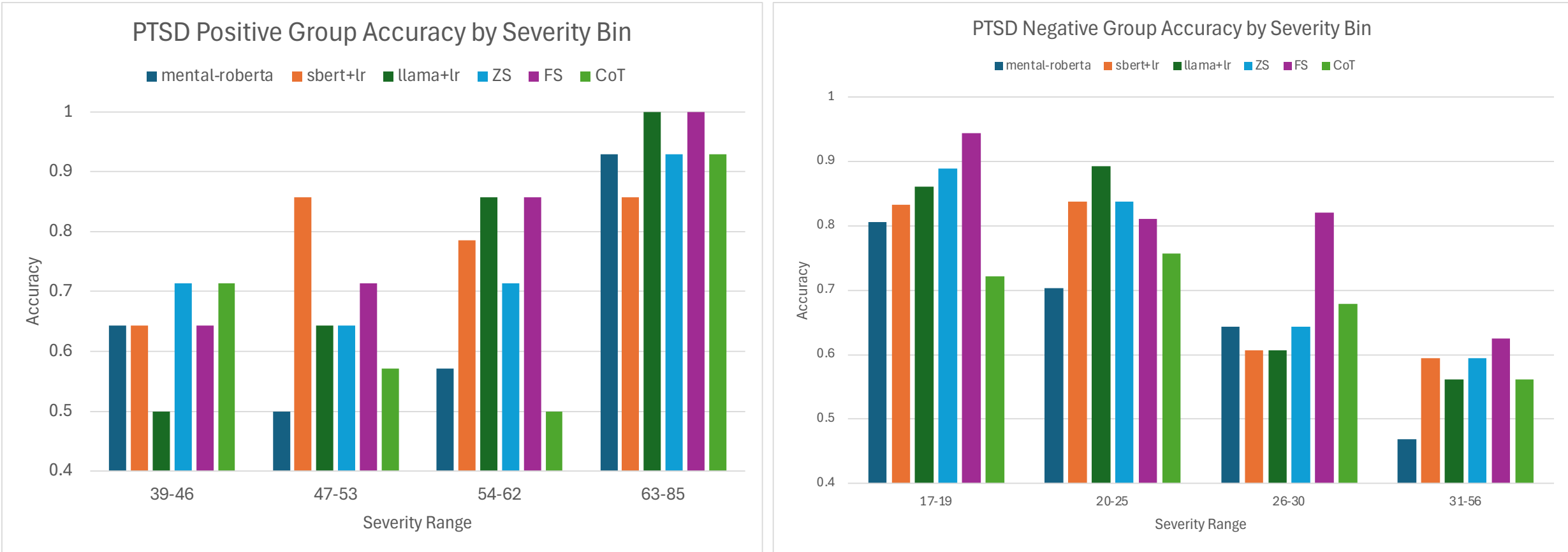

Figure 1. Accuracy of PTSD prediction by PCL-C severity score

### *3.4.2. Comorbidity Analysis*

We then evaluated the predictive accuracy for PTSD while considering the influence of depression comorbidity, with results shown in Table 4. A striking finding emerged when comparing model performance across depression comorbidity groups using AUPRC as the evaluation metric. For patients with comorbid depression, all models demonstrated remarkably high AUPRC scores: few-shot achieving 0.961, Mental-RoBERTa reaching 0.953, SentenceBERT + LR achieving 0.925, zero-shot reaching 0.915, LLaMA + LR achieving 0.879, and chain-of-thought reaching 0.859. In contrast, performance was substantially lower for PTSD classification among patients without depression, with AUPRC scores ranging from 0.381 to 0.574 across all models.

This performance pattern was consistent across all methodologies, suggesting that the observed performance disparity reflects a complex interplay between comorbidity status and symptom severity. Supporting this interpretation, our correlation analysis revealed strong associations between PTSD and depression severity scores (overall Spearman $\rho = 0.8426$, $p < 0.001$). Further analysis revealed that participants with depression exhibited significantly more severe PTSD symptoms, with a median PCL-C score of 59.5 compared to PTSD participants without comorbid depression (median = 49.5; Welch's t-test: $t=3.620$, $p<0.001$).

Table 4. AUPRC comparison of PTSD prediction in patient with/without depression

| AUPRC | Overall | patient with depression | Patient without depression |
|---|---|---|---|
| **Mental RoBERTa** | 0.675±0.084 | 0.953±0.048 | 0.521±0.171 |
| **SBERT + LR** | 0.758±0.128 | 0.925±0.073 | 0.574±0.137 |
| **LLaMA + LR** | 0.693±0.094 | 0.879±0.070 | 0.562±0.100 |
| **LLaMA zs** | 0.701 | 0.915 | 0.492 |
| **LLaMA fs** | 0.737 | 0.961 | 0.467 |
| **LLaMA cot** | 0.579 | 0.859 | 0.381 |

## 4. Discussion

Our findings demonstrate the potential of advanced NLP approaches for detecting PTSD from clinical interview transcripts, while also revealing important challenges and considerations for clinical implementation. The results highlight several key insights at the intersection of computational linguistics and mental health assessment.

Among supervised learning methods, embedding-based approaches demonstrated superior performance over end-to-end fine-tuning. The SentenceBERT embedding + logistic regression approach achieved the highest AUPRC performance (0.758±0.128), outperforming the best end-to-end model (Mental-RoBERTa: 0.675±0.084). This suggests that decoupling embedding generation from classification through a two-stage approach offers significant advantages for PTSD classification from clinical transcripts, particularly when working with limited labeled data. The effectiveness of logistic regression with balanced class weights demonstrates that sophisticated neural network architectures are not always necessary for clinical text classification when high-quality embeddings are available.

Within end-to-end fine-tuning approaches, domain-specific models (Mental-BERT and Mental-RoBERTa) substantially outperformed their general-domain counterparts (BERT-base and RoBERTa-base), underscoring the importance of domain adaptation in mental health applications. This finding aligns with prior work in clinical NLP that has shown domain-specific language representations better capture the nuanced linguistic patterns associated with psychological distress.[7, 26] The superior performance of mental health-specific models suggests that these pre-trained representations more effectively encode the subtle linguistic markers of PTSD.

Our exploration of more complex supervised learning approaches—including BERT/Mental BERT embeddings, deep feed-forward neural networks and complex pooling strategies—did not yield performance improvements over the embedding + logistic regression approach, suggesting that for clinical interview transcripts, simpler approaches may be more robust and generalizable.

Among instruction-tuned approaches, our exploration of LLM prompting strategies revealed that zero-shot classification using clinical criteria can achieve competitive performance without any training examples, highlighting the potential of instruction-tuned LLMs to leverage encoded linguistic information.[3] This finding has significant implications for low-resource scenarios where labeled clinical data is scarce or unavailable. The few-shot approach achieved the highest discriminative ability (AUROC: 0.875) among all instruction-tuned methods, suggesting that strategically selected examples can enhance LLM performance for clinical classification tasks, consistent with findings in other clinical NLP applications.[10]

Contrary to expectations, the chain-of-thought reasoning strategy showed lower performance despite its more structured analytical process. This pattern contrasts with findings in other domains where chain-of-thought reasoning typically enhances performance.[9] This could indicate that the linguistic markers of PTSD in transcripts are detected by LLMs more through implicit pattern recognition than through the explicit application of diagnostic criteria in a sequential reasoning process.

The substantial performance differences observed across symptom severity levels and comorbidity status represent particularly clinically relevant findings. Most models exhibited a clear bidirectional pattern: declining accuracy with increasing severity for PTSD-negative cases and improving accuracy with increasing severity for PTSD-positive cases. This severity-dependent performance mirrors challenges faced by human clinicians, where clear-cut cases are more readily identified than those with moderate or subclinical presentations.[27] This diagnostic complexity is further illustrated by the inconsistencies between PTSD diagnoses and patient-submitted PCL-C scores observed in the DAIC-WOZ dataset. Such inconsistencies typically arise because clinical interviews and self-report measures often diverge in moderate severity cases—clinicians may apply diagnostic thresholds differently than standardized cutoffs used in self-report scales, and patients' subjective experiences of symptoms may not align with clinically observable manifestations, creating a challenging middle ground where diagnostic agreement is lowest.

The markedly higher performance for patients with comorbid depression raises important considerations for clinical implementation. This enhanced model performance in comorbid cases likely reflects the more readily detectable transdiagnostic symptoms—such as emotional dysregulation, sleep disturbances, and cognitive difficulties—that represent cross-cutting dimensional constructs consistent with Research Domain Criteria (RDoC) frameworks.[28, 29] These shared symptom dimensions create stronger, more detectable linguistic signals when both conditions are present at higher severity levels. Rather than representing confounding between conditions, the substantial overlap in severe presentations indicates that comorbid cases manifest more pronounced psychological distress across multiple symptom domains, creating a more robust linguistic signature that facilitates computational detection.[30] This interpretation is strongly supported by the consistent pattern observed across all model architectures and prompting strategies, suggesting a fundamental linguistic phenomenon rather than a model-specific limitation.

This finding has two primary clinical implications. First, automated screening tools based on current approaches may be most effective as part of a staged screening process, where they are used to identify potential cases for further clinical assessment rather than as standalone diagnostic tools. This approach could provide value in places where mental health specialists are limited, serving as an efficient tool to flag cases requiring clinical attention. Second, these tools may require specific calibration for different patient populations, considering both symptom severity and comorbidity profiles. The development of more sophisticated models that can distinguish PTSD-specific linguistic markers from general indicators of psychological distress remains an important direction for future research.

## 5. Limitations and Future Directions

Several limitations should be considered when interpreting our findings. First, this study relied on a single dataset (DAIC-WOZ) with a relatively small sample size (189 participants), which may limit the generalizability of our findings. The limited sample size may explain why embedding-based approaches outperformed end-to-end fine-tuning approaches, as the latter require training the entire neural network and may be more prone to overfitting with insufficient training data. Second, the chunking and mean-pooling approach used to handle long transcripts may obscure the detection of temporal patterns that could be diagnostically relevant for PTSD, such as fragmented narratives when discussing traumatic events. While we explored alternative pooling strategies including attention-based and hierarchical long short-term memory pooling mechanisms, these did not improve performance over simple mean pooling. Third, while our approach incorporates DSM-5 criteria in prompting strategies, the linguistic patterns captured by computational models may not fully align with the clinical reasoning process used in traditional diagnostic approaches. Fourth, our few-shot prompting strategy used a simple selection method (highest and lowest PCL-C scores), which may not represent optimal example selection for enhancing model performance. More sophisticated example selection strategies based on linguistic diversity, diagnostic complexity, or active learning principles could potentially improve few-shot performance. Fifth, our evaluation was limited to a single LLM architecture (LLaMA 3.1), and performance may vary across different instruction-tuned models such as GPT-4, or other state-of-the-art LLMs. Finally, the DAIC-WOZ dataset, while valuable for its standardized interview format, may not fully represent the diversity of PTSD presentations across different trauma types, cultural backgrounds, and demographic groups. Evaluation on structured interview data may also limit generalizability to unstructured conversations. Future validation in diverse clinical settings and interview styles is essential to establish broader clinical utility.

Our work suggests several promising directions for future research. First, evaluation on larger, more diverse clinical datasets is needed to validate our findings and assess the scalability of embedding-based approaches. Second, multimodal approaches that combine linguistic analysis with audio and visual features could potentially enhance detection accuracy, particularly for patients who may not express their psychological distress linguistically. Third, developing more sophisticated sequence modeling approaches that can capture temporal dynamics while maintaining the efficiency of embedding-based methods represents an important technical challenge. Finally, future work should explore how these computational approaches can be integrated into clinical workflows to augment rather than replace clinician judgment, including developing interpretable models that provide clinicians with transparent reasoning for their predictions.

## 6. Conclusion

Our comprehensive evaluation of language modeling approaches for PTSD detection demonstrates both the potential and limitations of current computational methods. embedding-based methods, particularly SentenceBERT combined with logistic regression (AUPRC=0.758±0.128), outperform both domain-specific end-to-end fine-tuning and large language model prompting strategies. While domain-specific pre-training showed clear benefits over general models, and prompt-based approaches achieved competitive performance without requiring labeled data (few-shot AUROC=0.875), all methods demonstrated substantially higher performance for patients with comorbid depression and performed worse with moderate-severity cases, suggesting current approaches may detect general psychological distress rather than PTSD-specific markers. These findings provide practical insights for clinical implementation, highlighting the computational efficiency and clinical viability of embedding-based approaches while underscoring the need for more nuanced methods to address the significant challenges in PTSD recognition and facilitate earlier intervention for those affected by this widespread condition.

## Acknowledgements

This study was supported by National Institute of Mental Health U01MH135901. We would like to thank John Gratch, Jill Boberg, and their colleagues at the USC Institute for Creative Technologies for making this dataset available to the scientific community.

**References**

1.Affairs USDoV. How Common Is PTSD in Adults? National Center for PTSD2025 [Available from: https://www.ptsd.va.gov/understand/common/common_adults.asp.
2.Zammit S, Lewis C, Dawson S, et al. Undetected post-traumatic stress disorder in secondary-care mental health services: Systematic review. The British Journal of Psychiatry. 2018;212(1):11-8.
3.Stade EC, Stirman SW, Ungar LH, et al. Large language models could change the future of behavioral healthcare: a proposal for responsible development and evaluation. NPJ Mental Health Research. 2024;3(1):12.
4.Xu X, Yao B, Dong Y, et al. Mental-llm: Leveraging large language models for mental health prediction via online text data. Proceedings of the ACM on Interactive, Mobile, Wearable and Ubiquitous Technologies. 2024;8(1):1-32.
5.Omar M, Soffer S, Charney AW, et al. Applications of large language models in psychiatry: a systematic review. Frontiers in psychiatry. 2024;15:1422807.
6.Bartal A, Jagodnik KM, Chan SJ, Dekel S. AI and narrative embeddings detect PTSD following childbirth via birth stories. Scientific Reports. 2024;14(1):8336.
7.Ji S, Zhang T, Ansari L, et al. Mentalbert: Publicly available pretrained language models for mental healthcare. arXiv preprint arXiv:211015621. 2021.
8.Ogunleye B, Sharma H, Shobayo O. Sentiment Informed Sentence BERT-Ensemble Algorithm for Depression Detection. Big Data and Cognitive Computing. 2024;8(9):112.
9.Wei J, Wang X, Schuurmans D, et al. Chain-of-thought prompting elicits reasoning in large language models. Advances in neural information processing systems. 2022;35:24824-37.
10.Liu H, Zhang W, Xie J, et al. Few-shot learning for chronic disease management: Leveraging large language models and multi-prompt engineering with medical knowledge injection. arXiv preprint arXiv:240112988. 2024.
11.Gratch J, Artstein R, Lucas GM, et al., editors. The distress analysis interview corpus of human and computer interviews. LREC; 2014: Reykjavik.
12.DeVault D, Artstein R, Benn G, et al., editors. SimSensei Kiosk: A virtual human interviewer for healthcare decision support. Proceedings of the 2014 international conference on Autonomous agents and multi-agent systems; 2014.
13.Ringeval F, Schuller B, Valstar M, et al., editors. AVEC 2019 workshop and challenge: state-of-mind, detecting depression with AI, and cross-cultural affect recognition. Proceedings of the 9th International on Audio/visual Emotion Challenge and Workshop; 2019.
14.Burdisso S, Reyes-Ramírez E, Villatoro-Tello E, et al. DAIC-WOZ: On the Validity of Using the Therapist's prompts in Automatic Depression Detection from Clinical Interviews. arXiv preprint arXiv:240414463. 2024.
15.Devlin J, Chang M-W, Lee K, Toutanova K, editors. Bert: Pre-training of deep bidirectional transformers for language understanding. Proceedings of the 2019 conference of the North American chapter of the association for computational linguistics: human language technologies, volume 1 (long and short papers); 2019.
16.Liu Y, Ott M, Goyal N, et al. Roberta: A robustly optimized bert pretraining approach. arXiv preprint arXiv:190711692. 2019.
17.Loshchilov I, Hutter F. Decoupled weight decay regularization. arXiv preprint arXiv:171105101. 2017.

18.Kingma DP, Ba J. Adam: A method for stochastic optimization. arXiv preprint arXiv:14126980. 2014.
19.Cui Y, Jia M, Lin T-Y, et al., editors. Class-balanced loss based on effective number of samples. Proceedings of the IEEE/CVF conference on computer vision and pattern recognition; 2019.
20.Reimers N, Gurevych I. Sentence-bert: Sentence embeddings using siamese bert-networks. arXiv preprint arXiv:190810084. 2019.
21.Touvron H, Lavril T, Izacard G, et al. Llama: Open and efficient foundation language models. arXiv preprint arXiv:230213971. 2023.
22.Grattafiori A, Dubey A, Jauhri A, et al. The llama 3 herd of models. arXiv preprint arXiv:240721783. 2024.
23.Regier DA, Kuhl EA, Kupfer DJ. The DSM‐5: Classification and criteria changes. World psychiatry. 2013;12(2):92-8.
24.Andrykowski MA, Cordova MJ, Studts JL, Miller TW. Posttraumatic stress disorder after treatment for breast cancer: Prevalence of diagnosis and use of the PTSD Checklist—Civilian Version (PCL—C) as a screening instrument. Journal of consulting and clinical psychology. 1998;66(3):586.
25.Saito T, Rehmsmeier M. The precision-recall plot is more informative than the ROC plot when evaluating binary classifiers on imbalanced datasets. PloS one. 2015;10(3):e0118432.
26.Su C, Xu Z, Pathak J, Wang F. Deep learning in mental health outcome research: a scoping review. Translational psychiatry. 2020;10(1):116.
27.Palm KM, Strong DR, MacPherson L. Evaluating symptom expression as a function of a posttraumatic stress disorder severity. Journal of Anxiety Disorders. 2009;23(1):27-37.
28.Insel T, Cuthbert B, Garvey M, et al. Research domain criteria (RDoC): toward a new classification framework for research on mental disorders. American Psychiatric Association; 2010. p. 748-51.
29.Dalgleish T, Black M, Johnston D, Bevan A. Transdiagnostic approaches to mental health problems: Current status and future directions. Journal of consulting and clinical psychology. 2020;88(3):179.
30.Todorov G, Mayilvahanan K, Cain C, Cunha C. Context-and subgroup-specific language changes in individuals who develop PTSD after trauma. Frontiers in Psychology. 2020;11:989.